\RequirePackage{fix-cm}
\documentclass[smallextended]{svjour3}       % onecolumn (second format)
\smartqed  % flush right qed marks, i.e. at end of proof
\usepackage{graphicx}
\usepackage{amsmath}
\usepackage{amssymb}
\usepackage{color}
\usepackage{algorithmic}
\usepackage{algorithm}
\usepackage{subfigure}
\usepackage{array}
\usepackage{multirow,tabularx}
\usepackage[pagewise]{lineno}
\usepackage{multicol}
\usepackage{pdfsync}

\newcommand{\bfa}{{\textbf{a}}}
\newcommand{\bftau}{{\boldsymbol{\tau}}}
\newcommand{\bfomega}{{\boldsymbol{\omega}}}
\newcommand{\bfx}{{\textbf{x}}}
\newcommand{\bfv}{{\textbf{v}}}
\newcommand{\bfu}{{\textbf{u}}}
\newcommand{\bfw}{{\textbf{w}}}
\newcommand{\bfy}{{\textbf{y}}}

\newcommand{\bfmu}{{\boldsymbol{\mu}}}
\newcommand{\bfpi}{{\boldsymbol{\pi}}}

\newcommand{\bfphi}{{\boldsymbol{\phi}}}
\newcommand{\bfvarphi}{{\boldsymbol{\varphi}}}

\newcommand{\bfvartheta}{{\boldsymbol{\vartheta}}}

\begin{document}

\title{Cross Domain Adaptation by Learning Partially Shared Classifiers and Weighting Source Data Points in the Shared Subspaces}

\author{Hongqi Wang \and Anfeng Xu \and Shanshan Wang \and Sunny Chughtai}

\institute{Hongqi Wang, Anfeng Xu, and Shanshan Wang\at
School of Management of Harbin University of Science and Technology, Harbin 150000, China\\
\email{wanghongqi@yahoo.com, anfengxu150@yahoo.com}\\
Anfeng Xu is the corresponding author.
\and
Sunny Chughtai\at
Punjab University College of Information Technology (PUCIT), University of the Punjab, Lahore, Pakistan\\
\email{sunnychughtai313@yahoo.com}
}

\date{Received: date / Accepted: date}

\maketitle

\begin{abstract}
Transfer learning is a problem defined over two domains. These two domains share the same feature space and class label space, but have significantly different distributions. One domain has sufficient labels, named as source domain, and the other domain has few labels, named as target domain. The problem is to learn a effective classifier for the target domain. In this paper, we propose a novel transfer learning method for this problem by learning a partially shared classifier for the target domain, and weighting the source domain data points. We learn some shared subspaces for both the data points of the two domains, and a shared classifier in the shared subspaces. We hope that in the shared subspaces, the distributions of two domain can match each other well, and to match the distributions, we weight the source domain data points with different weighting factors. Moreover, we adapt the shared classifier to each domain by learning different adaptation functions. To learn the subspace transformation matrices, the classifier parameters, and the adaptation parameters, we build a objective function with weighted classification errors, parameter regularization, local reconstruction regularization, and distribution matching. This objective function is minimized by an iterative algorithm. Experiments show its effectiveness over benchmark data sets, including travel destination review data set, face expression data set, spam email data set, etc.
\keywords{Transfer Learning\and
Subspace Learning\and
Distribution Matching\and
Weighted Mean\and
Travel Destination Review}
\end{abstract}

%\linenumbers

\section{Introduction}
\label{sec:intro}

In this paper, we study the problem of transfer learning. This problem is defined over two domains, which is a source domain and a target domain. In the source domain, there are sufficient labeled data points. In the target domain, only a small number of data points are labeled. To learn a effective classifier for the target domain, we need the help of the source domain. However, the source domain and target domain have significantly different distributions, although they share the same feature and label space. Transfer learning, or named as cross domain adaptation, aims to fill the gap between the two domains to provide sufficient data for the target domain learning problem \cite{La2014807,Seera2014469,Seera20131,Yang2013541,Yang20121801,Tahmoresnezhad20161}.

\textbf{Example}. For example, in the problem of face expression, we want to train a classifiers for images of face to recognize the expression of this face. To this end, we can label the images of a training face image set of one individual, and when the images of anther individual appears, we can only label a few of his/her images, and training the classifier using the data of both individuals. The first individual is considered as a source domain, and the new individual is considered as target domain. The target domain has only a few labeled images while all the images of the source domain are labeled. Moreover, it is obvious that the distributions of these two individuals are different. Thus it is necessary to fill this gap for the learning from the two domains.

\subsection{Existing Works and Their Shortages}
\label{sec:existingworks}

There are plenty of works proposed to solve this problem, but they have some limitations. We discuss them as follows.

\begin{itemize}
\item Chen et al. \cite{Chen20131284} proposed a transfer learning framework for the adaptation of text mining models which is based on low-rank shared concept space. It minimize the gap between the distributions of the two domains, and meanwhile minimize the classification errors over the source domain. This method performs the domain adaptation over both the linear and kernel space of the features.
\item Chu et al. \cite{Chu6619295} proposed a transductive learning model, which is called selective transfer machine, for the problem of face expression recognition. This method tries to attenuate the person-specific biases of face expressions by personalizing the face expression classifiers. It solve this problem by jointly learning the source domain classifier and selecting the source domain data points for the learning problem. The selected source domain data points are assumed to be the most relevant data points to the target domain.
\item Ma et al. \cite{ma2014knowledge} proposed a transfer learning which can adapt knowledge from a source domain to a target domain, and can handle the features of the source and target domains which are partially different. The proposed framework minimizes the classification errors of the partially different features over both the source and target domains, while also hopes the classification results of the complete features of the target domain and the that of the features shared by the source domain can be consistent. Meanwhile the $\ell_{2,p}$ norm of the matrix of parameters of source and target domains are minimize so that the shared features of both domain can be jointly selected.
\item Xiao and Guo \cite{Xiao201554} proposed a transfer learning method to handle the domain adaptation problem with completed feature spaces for the source and target domains. This method is based on the learning of classifiers of the source domain and the mapping of the target domain data points to the source domain. The mapping is conducted by the kernel matching of the kernel matrix of the two domains. The mapping of the kernel matrices is based on the Hilbert Schmidt Independence Criterion.

\item Li et al. \cite{Li20141134} proposed method for the heterogeneous domain adaptation problem. In this transfer learning problem, the features from the source and target domains can also be different. To handle different feature spaces, two different projection matrices are applied to them, so that they can be mapped to a shared subspace. Moreover, a new features mapping method is proposed for each domain, and the data points are augmented with the transformed features and the original features. Furthermore, the support vector machine framework is employed to learn the transformation matrices and the classifier parameters.
\end{itemize}

The \textbf{shortages} of these works paper are of three folds.

\begin{itemize}
\item Some methods ignore the label information of the target domain, which is critical for the learning of the target domain. Works of Chen et al. \cite{Chen20131284} and Chu et al. \cite{Chu6619295} do not use the labels of the target domain data points. Without using the target domain label information, only minimizing the gap of distributions between the two domains cannot guarantee the learned classifier not over-fitting to the source domain.
\item Some works also ignore the local connection of the data points of both source and target domain. The learning is performed over the data points independently. However, researches of manifold learning showed that it is important to explore the local connections between the data points when the classifier is learned. Among all the methods mentioned above, only the work of Xiao and Guo \cite{Xiao201554} uses the local connection information to regularize the learning of the classifiers.
\item Some works ignore the differences of the source domain data points for the learning problem of target domain. It treats all the source domain data points evenly. However, we observed that actually different source domain data points paly different roles in the learning of target domain classifier. Only the work of Chu et al. \cite{Chu6619295} select the source domain data points for the learning of the target domain classifier, while all the other methods treat them source domain data points equally.
\end{itemize}

\subsection{The Contributions}

In this paper, we propose a novel transfer learning problem. This method is aimed to solve the problems of existing transfer learning methods. The learning framework and model is summarized as follows,

\begin{itemize}
\item To solve the problem of ignoring the labels of the target domain, we proposed to minimize the classification errors of the data points of both the source and target domains. The classification errors are measured by the loss functions such as logistic loss and hinge loss.
\item To solve the problem of ignoring the differences of importance of different source domain data points, we propose to weight them with different weighting factors. The weight factors are used in two different scenes. The first scene is the matching of the distributions of the two domains. We propose to use the weighted mean of the source domain to match the original mean of the target domain. In this way, the source domain data points which match better to the target domain can receive larger weighting factors. In the second scene, the loss terms of these data points with larger weighting factors are also assigned with larger weights in the loss function.
\item To solve the problem of ignoring the local connection information, we propose to use local reconstruction information to regularize the learning of the weights of the source domain data points, and the classifier of the target domain. The local reconstruction coefficients are learned from the original features.
\item Moreover, an inessential contribution of this work is a novel cross domain classification framework to implement the above three thoughts. To gap the fill of the two domains, we propose to map the data of two domains to some shared subspaces, and then design a shared classifier in this space. In these shared subspaces, we match the distributions by using the weighting factors of the source domain data points. Moreover, we propose to adapt this shared classifier to source and target domains by addling adaptation functions of the original spaces. The final classifiers are actually partially shared classifiers. The loss functions are defined based on these partially shared classifiers, and the local reconstruction regularization is also performed over the partially shared classifier.
\end{itemize}

The modeling of the learning problem is based on the three thoughts mentioned above, and each thought is corresponding to a term of the objective function. The learning problem is constructed by minimizing the objective function with regard to the parameters of the mapping matrix, the shared classifier parameter and the adaptation parameters. Moreover, the weighting factors should also be considered as variables to learn. We design an iterative learning algorithm to solve this problem.

\subsection{Organization of the Rest Parts}

The paper is organized as follows. In section \ref{sec:model}, we introduce the model of the proposed data representation and classification method, and the learning method of the parameters of this model is introduced in section \ref{sec:learning}. In section \ref{sec:exp}, the proposed algorithm is evaluated over some benchmark data sets. In section \ref{sec:conclu}, the conclusion is given, and in section \ref{sec:future}, the future work is summarized.

\section{Transfer Representation and Classification Model}
\label{sec:model}

\subsection{Shared Subspace Representation}

Suppose we have a source domain and a target domain, and each domain has a training set. The source domain training set is give as a set of $n_1$ labeled data points, $\mathcal{S} = \{(\bfx_1^s, y_1^s), \cdots, (\bfx_{n_1}^s, y_{n_1}^s)\}$, where $\bfx_i^s\in \mathbb{R}^m$ is the feature vector of $m$ dimensions of the $i$-th data point, and $y_i^s\in \{+1,-1\}$ is the binary label of the $i$-th data point. The target domain training set is given as a set of $n_2$ partially labeled data points, $\mathcal{T} = \{(\bfx_1^t,y_1^t), \cdots, (\bfx_{n_3}^t,y_{n_3}^t), \bfx_{n_3+1}^t,\cdots, \bfx_{n_2}^t\}$, where $\bfx_j^t\in \mathbb{R}^m$ is the feature vector of the $j$-th data point, and $y_j^t\in \{+1,-1\}$ is the binary label of the $j$-th data point. In the training set of the target domain, the first $n_3$ data points are labeled, while the remaining $n_2-n_3$ data points are not labeled. To represent the data points from both the source and target domains, we proposed to map them to $r$ shared subspaces by a transformation matrix $\Theta\in \mathbb{R}^{r\times m}$, where $r< m$ is the number of subspaces. Given a data point $\bfx$, we use the transformation matrix to project it to a lower dimensional space, and the feature vector in this space $\bfy$ is obtained as,

\begin{equation}
\label{equ:y}
\begin{aligned}
\bfy = \Theta \bfx.
\end{aligned}
\end{equation}
Please note that this subspace is shared by both the source and target domain, thus it makes it possible to learn a shared classifier for the two domains. To this end, we designed a linear classifier and apply it to a feature vector, $\bfy$, in the subspace which is obtained from (\ref{equ:y}),

\begin{equation}
\label{equ:g}
\begin{aligned}
g(\bfx) = \bfw^\top \bfy = \bfw^\top \Theta \bfx,
\end{aligned}
\end{equation}
where $\bfw \in \mathbb{R}^r$ is the parameter vector of the shared classifier $g$.

\subsection{Filling Gap Between Source and Target Domains}

Although the shared subspace and classifier seems to work well for both domains, there is a gap between these two domains. Usually the two domains have significant different distributions in the original feature space, and simply mapping the data of two domains to shared subspaces will not automatically fill this gap. To solve this problem, we propose two possible solutions.

\subsubsection{Partially Shared Classifiers}
The first solution is to design the classifiers of source and target domain by adapting the shared classifier to the two domains respectively. The adaptation is implemented by adding a linear function over the original feature vectors. For a source domain data point, $\bfx^s$, the classifier is given as follows,

\begin{equation}
\label{equ:fs}
\begin{aligned}
f(\bfx^s) = g(\bfx^s) +\Delta_s(\bfx^s) = \bfw^\top \Theta \bfx^s + \bfu^\top \bfx^s,
\end{aligned}
\end{equation}
where $\Delta_s(\bfx^s) = \bfu^\top \bfx^s $ is the source domain adaption function and $\bfu \in \mathbb{R}^m$ is the parameter vector of the adaptation function. For a target domain data point, $\bfx^t$, the classifier is obtained as the combination of $g$ and a target domain function,

\begin{equation}
\label{equ:ht}
\begin{aligned}
h(\bfx^t) = g(\bfx^t) +\Delta_t(\bfx^t) = \bfw^\top \Theta \bfx^t + \bfv^\top \bfx^t,
\end{aligned}
\end{equation}
where $\Delta_t(\bfx^t) = \bfv^\top \bfx^t$ is the target domain adaptation function, and $\bfv\in \mathbb{R}^m$ is its parameter vector. In this way, the source and target domain has a shared classifier, $g$, and two independent adaptation functions, $\Delta_s$ and $\Delta_t$. Using the subspace projection shared by both two domains and its corresponding classifier $g$, the gap between the two domains are somehow filled. Meanwhile, we also respect the difference between the two domains by adapting the shared classifier to two different domains respectively.
%This learning framework is summarized in Figure . From this figure, we can see that the source and target domain classifiers are both composed of a shared classifier $g$ which is based on some shared subspaces, and domain-specific adaptation terms. Thus we call our classification framework partially shared classifiers.
In this classification framework, we have several parameters to learn, including $\Theta$, $\bfw$, $\bfu$, and $\bfv$. To estimate these parameters, we propose a novel learning framework with an objective function and an optimization method.

\subsubsection{Distribution Matching in the Subspaces by Weighting the Source Domain Data Points}

The second solution to fill the gap is to match the distributions of the two domains in the subspaces. Since the subspaces are shared by both the domains, we hope that in these subspaces, the gap between distributions of source and target domains can be filled. To this end, we proposed to represent the distributions of subspaces of source and target domains as the mean of the vectors of their data points in the subspaces,

\begin{equation}
\label{equ:mu}
\begin{aligned}
&\bfmu_s = \frac{1}{n_1} \sum_{i=1}^{n_1} \Theta \bfx_i^s,~and\\
&\bfmu_t = \frac{1}{n_2} \sum_{j=1}^{n_2} \Theta \bfx_j^t
\end{aligned}
\end{equation}
where $\bfmu_s$ and $\bfmu_t$ are the mean vectors of source and target domains respectively. Moreover, we argue that actually different source domain data points plays different roles in the matching of distributions of two domains. It is not suitable to treat all the source domain data points evenly when they are matched to the target domain. Some source domain data points are more important to the learning of target domain classifier, and they matches to the target domain better than the other source domain data points. Thus we propose to weight the source domain data points with different weights, so that the important data points can obtain larger weights than other data points. The nonnegative weight of the $i$-th source domain data points is defined as $\pi_i$, and the vector of weights is defined as $\bfpi = [\pi_1, \cdots, \pi_{n_1}]$. With these weights, we refine the mean vector of the source domain as follows,

\begin{equation}
\label{equ:mus}
\begin{aligned}
&\bfmu_s^\bfpi = \frac{1}{n_1} \sum_{i=1}^{n_1} \Theta \bfx_i^s \pi_i.
\end{aligned}
\end{equation}
In this refined mean, each data point in the subspaces are weighted by a factor $\pi_i$. To match the refined distribution mean of the source domain and the target domain mean, we propose to minimize the squared $\ell_2$ norm distance between them with regard to both $\Theta$ and $\pi$,

\begin{equation}
\label{equ:match}
\begin{aligned}
\min_{\Theta,\pi} ~~ & \frac{1}{2} \| \bfmu_s^\bfpi - \bfmu_t\|_2^2\\
& = \frac{1}{2} \left \| \frac{1}{n_1} \sum_{i=1}^{n_1} \Theta \bfx_i^s \pi_i -
\frac{1}{n_2} \sum_{j=1}^{n_2} \Theta \bfx_j^t \right \|_2^2.
\end{aligned}
\end{equation}
By solving this problem, we argue that a good combination of $\Theta$ and $\pi$ should make the data distributions of source domain and target domain match each other well, i.e., make the squared $\ell_2$ norm distance in (\ref{equ:match}) minimum. In this problem, we need to approximate the weighting factors in $\bfpi$.

\section{Parameter Estimation Method}
\label{sec:learning}

In this section, we build a joint learning framework to learn the parameters of our classification methods. We will first discuss the objective function of the learning problem, and then minimize the objective function with regard to the variables.

\subsection{Objective function}

The objective function of the learning framework is given as follows,

\begin{equation}
\label{equ:objective}
\begin{aligned}
\mathcal{O}&(\Theta,\bfw,\bfu,\bfv,\bfpi)
=
\sum_{i=1}^{n_1} L(y_i^s, f(\bfx^s_i)) \pi_i + \sum_{j=1}^{n_3} L(y_i^t, h(\bfx^t_j))\\
&+\frac{C_1}{2} \left ( \|\bfu\|_2^2 + \|\bfv\|_2^2 \right )\\
&+C_2 \left ( \sum_{i=1}^{n_1}  \left \|\pi_i - \sum_{k\in \mathcal{N}^s_i} \omega_{ik}^s \pi_k \right \|_2^2 +
\sum_{j=1}^{n_2}  \left \|h(\bfx^t_j) - \sum_{k'\in \mathcal{N}^t_j} \omega_{jk'}^t h(\bfx^t_{k'}) \right \|_2^2
 \right )\\
&+\frac{C_3 }{2} \left \| \frac{1}{n_1} \sum_{i=1}^{n_1} \Theta \bfx_i^s \pi_i -
\frac{1}{n_2} \sum_{j=1}^{n_2} \Theta \bfx_j^t \right \|_2^2.
\end{aligned}
\end{equation}
In this objective function, we have four types of terms. We discuss them as follows in details.

\begin{itemize}
\item \textbf{Losses of classification errors.} The first two terms of the objective function are the classification losses of the data points of source and target domains. In both terms, $L(y,f(\bfx))$ is the loss function of a data point, where $\bfx$ is the input, $f$ is the classifier, and $y$ is the true class label. We discuss the following three types of loss functions,

\begin{enumerate}
\item Hinge loss: $L(y,f(\bfx)) = \max(0, 1-yf(\bfx))$,
\item Logistic loss: $L(y,f(\bfx)) = \log \left ( 1 + \exp (- yf(\bfx))\right )$, and
\item Exponential loss: $L(y,f(\bfx)) = \exp (- yf(\bfx))$.
\end{enumerate}

The first term, $\sum_{i=1}^{n_1} L(y_i^s, f(\bfx^s_i)) \pi_i$ is the weighted summation of the losses of classification errors the source domain data points. Each loss is weighted by its corresponding weighting factor of the distribution matching. The motive is that we observed that if a source domain data point is important for the matching of source and target domain matching in the subspaces, it is also important for the learning of classifier of target domain. Using the same weighting factor to regularize the learning of the source domain classifier also regularize the learning of the subspaces and shared classifier, which is critical for the target domain.

The second term, $\sum_{j=1}^{n_3} L(y_i^t, h(\bfx^t_j))$, is the losses of classification of the labeled target domain data points. Only the first $n_3$ labeled data points are considered in this loss function because only their labels are available.

\item \textbf{Adaptation function parameter regularization.} We further propose to regularize the parameter vectors of both the source and target domain adaptation functions to avoid the over-fitting to different domains. The motive for this is that since the classifiers are adapted from a shared subspace classifier, we do not want to make the adaptation too complex and over-fitted to the training sets of two different domains, so that the gap will not be enlarged but filled. To this end, we argue that the adaption parameters should be as simple as possible. To measure the complexities, we use the squared $\ell_2$ norms of the parameters of both $\Delta_s$ and $\Delta_t$, $\bfu$ and $\bfv$. Thus the third term of (\ref{equ:objective}), $\frac{C_1}{2} \left ( \|\bfu\|_2^2 + \|\bfv\|_2^2 \right )$, is minimized. $C_1$ is the weight of this term in the objective function.

\item \textbf{Neighborhood reconstruction regularization of $\pi$ and target domain classifier.} We argue the if the neighbors of a source domain data points are receiving large weights, itself should also receive large weights. We denote the neighbor set of a source data point $\bfx^s_i$ as $\mathcal{N}_i^s$. To impose our argument, we propose to reconstruct each data point $\bfx_i^s$ for its neighbors in $\mathcal{N}_i^s$, and further use the reconstruction coefficients to regularize the learning of $\pi$. The reconstruction coefficients are solved as the following minimization problem,

\begin{equation}
\label{equ:reconstruction}
\begin{aligned}
\min_{\omega_{ik},k\in \mathcal{N}_i^s}
~&\left \|\bfx_i^s - \sum_{k\in \mathcal{N}_i^s} \omega_{ik}^s \bfx_k^s\right \|_2^2\\
s.t.~&\sum_{k\in \mathcal{N}_i^s} \omega_{ik}^s  =1,\omega_{ik}^s \geq 0,\forall~k\in \mathcal{N}_i^s
\end{aligned}
\end{equation}
where $\omega_{ik}^s ,k\in \mathcal{N}_i^s$ are the coefficients for reconstruction of $\bfx_i^s$ from the neighbors in $\mathcal{N}_i^s$, and $\left \|\bfx_i^s - \sum_{k\in \mathcal{N}_i^s} \omega_{ik} \bfx_k^s\right \|_2^2$ is the reconstruction error measure. Using the reconstruction coefficients, we regularize the learning of the weighting factors, by minimizing the reconstruction error in the space of $\pi$, by minimizing $\left \|\pi_i - \sum_{k\in \mathcal{N}^s_i} \omega_{ik}^s \pi_k \right \|_2^2$. In this case, if $\omega_{ik}^s$ is large, it means $\bfx_k^s$ contribute significantly to the reconstruction of $\bfx_i^s$, and the similarity between $\bfx_i^s$ and $\bfx_k^s$ is large. Thus we also hope $\pi_i$ and $\pi_k$ can be similar to each other.

Similarly, we also regularize the learning of the target domain classifier by the neighborhood reconstruction coefficients. The motive is to use the unlabeled data points of the target domain. These data points are not used in the classification error part of the objective. They are used to match the distributions of the two domains, however, the distribution matching does not consider the label information. Thus we propose to propagate the label information by the neighborhood reconstruction regularization. In the third term of the objective of (\ref{equ:objective}), the reconstruction error of the classification responses are also minimized as  $\sum_{j=1}^{n_2}  \left \|h(\bfx^t_j) - \sum_{k'\in \mathcal{N}^t_j} \omega^t_{jk'} h(\bfx^t_{k'}) \right \|_2^2$. By minimizing this term, we hope that the neighboring target domain data points also have neighboring classification results. In this term, $\mathcal{N}^t_j$ is the set of neighboring target domain data points of $\bfx_j^t$, and $\omega_{jk'}^t, k'\in \mathcal{N}^t_j$ are the reconstruction coefficient of $\bfx_j^t$ from $\mathcal{N}^t_j$.

\item The last term is the target domain distribution matching term. We have discussed this term in (\ref{equ:match}). $C_3$ is the weight for this term.

\end{itemize}

The learning problem is to minimize this objective function with regard to the parameters of $\Theta$, $\bfw$, $\bfu$, $\bfu$ and $\bfpi$,

\begin{equation}
\label{equ:minim}
\begin{aligned}
\min_{\Theta,\bfw,\bfu,\bfv,\bfpi}~&
\mathcal{O}(\Theta,\bfw,\bfu,\bfv,\bfpi),\\
s.t.~&\Theta \Theta^\top = I_r,\\
&\bf0 \leq\bfpi \leq \delta\bf1, ~and~\bfpi^\top \bf1 = n_1.
\end{aligned}
\end{equation}
In this minimization problem, we impose the constraint $\Theta \Theta^\top = I_r$, where $I_r$ is a $r\times r$ identity matrix, so that the subspace transformation matrix $\Theta$ is orthogonal. Moreover, we impose a lower bound for $\bfpi$, $\bf0$ which is a vector of all zeros, and a upper bound of $\bfpi$, $\delta\bf1$, where $\bf1$ is a vector of all ones. We also propose an additional constraint to $\bfpi$, so that the summation of all the elements of $\bfpi$ is $n_1$. The motive of this constraint is that for the original calculation of $\bfmu_s$ of (\ref{equ:mu}), actually we set the weight of each data point to one, and the summation of the weights is $n_1$. This makes it comparable to the scale of $\bfmu_t$. To maintain this property when we use $\bfmu_s^\bfpi$ to replace $\bfmu_s$, we still requires that the summation of the weights in $\bfpi$ to be one.

\subsection{Solution}

In this section, we discuss how to solve the minimization problem in (\ref{equ:minim}), and design an iterative algorithm based on the solutions. To make the problem easier to solve, we rewrite the source domain and target domain classifiers as a linear function of the input feature vectors,

\begin{equation}
\label{equ:fs1}
\begin{aligned}
f(\bfx^s)
&= \bfw^\top  \bfx^s + \bfu^\top \bfx^s\\
&= \bfphi^\top \bfx^s, ~where\\
\bfphi&=\Theta^\top\bfw + \bfu,~and\\
h(\bfx^t) &= \bfw^\top \Theta \bfx^t + \bfv^\top \bfx^t, \\
&= \bfvarphi^\top \bfx^t, ~where\\
\bfvarphi&= \Theta^\top\bfw+\bfv.
\end{aligned}
\end{equation}
In this way, we present the adaptation parameter vectors  $\bfu$ and $\bfv$ as functions of $\Theta$, $\bfw$, and $\bfphi$ or $\bfvarphi$,

\begin{equation}
\label{equ:uv}
\begin{aligned}
\bfu &= \bfphi - \Theta^\top\bfw,~and\\
\bfv &= \bfvarphi - \Theta^\top\bfw.
\end{aligned}
\end{equation}
Substituting (\ref{equ:fs1}) and (\ref{equ:uv}) back to (\ref{equ:objective}) and (\ref{equ:minim}), we have the following minimization problem,

\begin{equation}
\label{equ:minim1}
\begin{aligned}
\min_{\Theta,\bfw,\bfphi,\bfvarphi,\bfpi}~&
\sum_{i=1}^{n_1} L(y_i^s, \bfphi^\top \bfx^s) \pi_i + \sum_{j=1}^{n_3} L(y_i^t, \bfvarphi^\top \bfx^t)\\
&+\frac{C_1}{2} \left ( \|\bfphi - \Theta^\top\bfw\|_2^2 + \|\bfvarphi - \Theta^\top\bfw\|_2^2 \right )\\
&+C_2 \left ( \sum_{i=1}^{n_1}  \left \|\pi_i - \sum_{k\in \mathcal{N}^s_i} \omega_{ik}^s \pi_k \right \|_2^2 +
\sum_{j=1}^{n_2}  \left \|\bfvarphi^\top \bfx^t_j - \sum_{k'\in \mathcal{N}^t_j} \omega_{jk'}^t \bfvarphi^\top \bfx^t_{k'} \right \|_2^2
 \right )\\
&+\frac{C_3 }{2} \left \| \frac{1}{n_1} \sum_{i=1}^{n_1} \Theta \bfx_i^s \pi_i -
\frac{1}{n_2} \sum_{j=1}^{n_2} \Theta \bfx_j^t \right \|_2^2,\\
s.t.~&
\Theta \Theta^\top = I_r,\\
&\bf0 \leq\bfpi \leq \delta\bf1, ~and~\bfpi^\top \bf1 = n_1.
\end{aligned}
\end{equation}
To solve this problem, we use the iterative optimization method. We propose to update the parameters one by one in an iteration of an iterative algorithm. When on parameter is updated, the other parameters are fixed. In the following subsections, we will discuss how to update the parameters one by one.

\subsubsection{Updating $\Theta$ and $\bfw$}

When we update $\Theta$ and $\bfw$, we fix the other parameters and obtain the following minimization problem,

\begin{equation}
\label{equ:minTheta}
\begin{aligned}
\min_{\Theta,\bfw}~&
\frac{C_1}{2} \left ( \|\bfphi - \Theta^\top\bfw\|_2^2 + \|\bfvarphi - \Theta^\top\bfw\|_2^2 \right )\\
&+\frac{C_3 }{2} \left \| \frac{1}{n_1} \sum_{i=1}^{n_1} \Theta \bfx_i^s \pi_i -
\frac{1}{n_2} \sum_{j=1}^{n_2} \Theta \bfx_j^t \right \|_2^2,\\
s.t.~&
\Theta \Theta^\top = I_r.
\end{aligned}
\end{equation}
Please note that in the objective of this minimization problem, the terms which do not contains $\Theta$ and $\bfw$ have been removed. To minimize this objective with regard to $\bfw$, we set is derivative with regard to $\bfw$ to zero, and obtain the optimal solution of $\bfw$ as follows,

\begin{equation}
\label{equ:woptimal}
\begin{aligned}
&C_2\Theta \left [ (\bfphi - \Theta^\top\bfw) + (\bfvarphi - \Theta^\top \bfw)  \right ] = 0\\
&\Rightarrow
C_2\Theta \left [ (\bfphi + \bfvarphi) - 2\Theta^\top\bfw  \right ] = 0\\
&\Rightarrow
C_2\Theta (\bfphi + \bfvarphi)  = C_2 2\bfw \\
&\Rightarrow
\bfw = \frac{1}{2}\Theta (\bfphi + \bfvarphi).
\end{aligned}
\end{equation}
Substituting $\bfw$ of (\ref{equ:woptimal}) back to (\ref{equ:minTheta}), we have

\begin{equation}
\label{equ:minTheta}
\begin{aligned}
\min_{\Theta}~
&
\frac{C_1}{2} \left ( \left \|\bfphi - \frac{1}{2} \Theta^\top
\Theta (\bfphi + \bfvarphi) \right \|_2^2 + \left \|\bfvarphi - \frac{1}{2}\Theta^\top\Theta (\bfphi + \bfvarphi)\right \|_2^2 \right )\\
&+\frac{C_3 }{2} \left \| \frac{1}{n_1} \sum_{i=1}^{n_1} \Theta \bfx_i^s \pi_i -
\frac{1}{n_2} \sum_{j=1}^{n_2} \Theta \bfx_j^t \right \|_2^2,\\
&=
\frac{C_1}{2} \left ( \left \|\bfphi - \frac{1}{2} \Theta^\top
\Theta (\bfphi + \bfvarphi) \right \|_2^2 + \left \|\bfvarphi - \frac{1}{2}\Theta^\top\Theta (\bfphi + \bfvarphi)\right \|_2^2 \right )\\
&+\frac{C_3 }{2} \left \| \frac{1}{n_1} \sum_{i=1}^{n_1} \Theta \bfx_i^s \pi_i -
\frac{1}{n_2} \sum_{j=1}^{n_2} \Theta \bfx_j^t \right \|_2^2,\\
&=
\frac{C_1}{2} \left [
\bfphi^\top \bfphi + \bfvarphi^\top \bfvarphi - \frac{1}{2} Tr \left ( \Theta (\bfphi+\bfvarphi)(\bfphi+\bfvarphi)^\top \Theta^\top
\right )\right ]\\
&+\frac{C_3 }{2} Tr \left (\Theta
\left ( \frac{1}{n_1} \sum_{i=1}^{n_1}  \bfx_i^s \pi_i -
\frac{1}{n_2} \sum_{j=1}^{n_2}  \bfx_j^t \right )
\left ( \frac{1}{n_1} \sum_{i=1}^{n_1}  \bfx_i^s \pi_i -
\frac{1}{n_2} \sum_{j=1}^{n_2}  \bfx_j^t \right ) ^\top
\Theta^\top \right ),\\
& = \frac{C_1}{2}\left ( \bfphi^\top \bfphi + \bfvarphi^\top \bfvarphi\right ) + Tr(\Theta \Phi \Theta^\top),\\
s.t.~&
\Theta \Theta^\top = I_r,
\end{aligned}
\end{equation}
where $Tr(X)$ is the trace of a matrix $X$, and

\begin{equation}
\label{equ:Phi}
\begin{aligned}
\Phi =
&-\frac{C_1}{4}  (\bfphi+\bfvarphi)(\bfphi+\bfvarphi)^\top
%\vphantom{}
\\
&
+\frac{C_3}{2}
\left ( \frac{1}{n_1} \sum_{i=1}^{n_1}  \bfx_i^s \pi_i -
\frac{1}{n_2} \sum_{j=1}^{n_2}  \bfx_j^t \right )
\left ( \frac{1}{n_1} \sum_{i=1}^{n_1}  \bfx_i^s \pi_i -
\frac{1}{n_2} \sum_{j=1}^{n_2}  \bfx_j^t \right ) ^\top.
\end{aligned}
\end{equation}
The first term of (\ref{equ:minTheta}) is irrelevant to $\Theta$, thus we can remove it from the objective. The problem in (\ref{equ:minTheta}) is further turned to

\begin{equation}
\label{equ:minTheta1}
\begin{aligned}
\min_{\Theta}~
&
Tr(\Theta \Phi \Theta^\top)\\
s.t.~&
\Theta \Theta^\top = I_r.
\end{aligned}
\end{equation}
To solve this problem, we should decompose $\Phi$ by the eigen-decomposition,

\begin{equation}
\label{equ:Phi2}
\begin{aligned}
\Phi = \Psi \Lambda \Psi^\top
\end{aligned}
\end{equation}
where $\Lambda$ is the diagonal matrix of eigenvalues, and the rows of $\Psi$ contain the rows of eigenvectors  corresponding to the eigenvalues. We pick up the $r$ rows corresponding to the largest $r$ eigenvalues from $\Psi$, and the matrix of these $r$ rows are the optimal solution of  $\Phi$.

\subsubsection{Updating $\bfphi$ and $\bfvarphi$}

In this step, we fixe the other parameters and consider only $\bfphi$ and $\bfvarphi$. When other parameters are fixed and the terms which do not contain $\bfphi$ and $\bfvarphi$ are removed, the minimization problem in (\ref{equ:minim1}) is reduced to the following problem,

\begin{equation}
\label{equ:minimuv}
\begin{aligned}
\min_{\bfphi,\bfvarphi}~&  \mathcal{Q}(\bfphi,\bfvarphi) =
\sum_{i=1}^{n_1} L(y_i^s, \bfphi^\top \bfx^s) \pi_i + \sum_{j=1}^{n_3} L(y_i^t, \bfvarphi^\top \bfx^t)\\
&+\frac{C_1}{2} \left ( \|\bfphi - \Theta^\top\bfw\|_2^2 + \|\bfvarphi - \Theta^\top\bfw\|_2^2 \right )\\
&+C_2
\sum_{j=1}^{n_2}  \left \|\bfvarphi^\top \bfx^t_j - \sum_{k'\in \mathcal{N}^t_j} \omega_{jk'}^t \bfvarphi^\top \bfx^t_{k'} \right \|_2^2.
\end{aligned}
\end{equation}
To solve this problem, we use the coordinate descent algorithm. This algorithm can minimize a function of multiple variables. When one variable is updated, we update the variable toward the direction of the sub-gradient, while other variable are fixed. The sub-gradient functions of $\mathcal{Q}$ with regard to $\bfphi$ and $\bfvarphi$ are as follows,

\begin{equation}
\label{equ:suggra}
\begin{aligned}
\nabla\mathcal{Q}_{\bfphi}=
&\sum_{i=1}^{n_1} \nabla L_\bfphi(y_i^s, \bfphi^\top \bfx^s) \pi_i +C_1 (\bfphi - \Theta^\top\bfw),~and\\
\nabla\mathcal{Q}_{\bfvarphi}=
&\sum_{j=1}^{n_3} \nabla L_\bfvarphi(y_i^t, \bfvarphi^\top \bfx^t)+C_1 (\bfvarphi - \Theta^\top\bfw)\\
&+2 C_2
\sum_{j=1}^{n_2}   \left ( \bfx^t_j - \sum_{k'\in \mathcal{N}^t_j} \omega_{jk'}^t \bfx^t_{k'} \right )
\left ( \bfx^t_j - \sum_{k'\in \mathcal{N}^t_j} \omega_{jk'}^t \bfx^t_{k'} \right )^\top \bfvarphi.
\end{aligned}
\end{equation}
The updating rules are given as follows,

\begin{equation}
\label{equ:updateuv}
\begin{aligned}
\bfphi\leftarrow \bfphi - \rho \nabla\mathcal{Q}_{\bfphi}, ~and~
\bfvarphi\leftarrow \bfvarphi - \rho \nabla\mathcal{Q}_{\bfvarphi},
\end{aligned}
\end{equation}
where $\rho$ is the step of descent.

\subsubsection{Updating $\bfpi$}

In this step, we update $\bfpi$ and fix other parameters. To this end, we have the following minimization problem,

\begin{equation}
\label{equ:minimpi}
\begin{aligned}
\min_{\bfpi}~&
\sum_{i=1}^{n_1} L(y_i^s, \bfphi^\top \bfx^s) \pi_i
+C_2 \sum_{i=1}^{n_1}  \left \|\pi_i - \sum_{k\in \mathcal{N}^s_i} \omega_{ik}^s \pi_k \right \|_2^2\\
&+\frac{C_3 }{2} \left \| \frac{1}{n_1} \sum_{i=1}^{n_1} \Theta \bfx_i^s \pi_i -
\frac{1}{n_2} \sum_{j=1}^{n_2} \Theta \bfx_j^t \right \|_2^2,\\
s.t.~&\bf0 \leq\bfpi \leq \delta\bf1, ~and~\bfpi^\top \bf1 = n_1.
\end{aligned}
\end{equation}
The first term of the objective can be rewritten as

\begin{equation}
\label{equ:linear}
\begin{aligned}
&\sum_{i=1}^{n_1} L(y_i^s, \bfphi^\top \bfx^s) \pi_i = \bftau^\top \bfpi, \\
&where~\bftau=[\tau_1,\cdots,\tau_{n_1}]^\top,~and~\tau_i = L(y_i^s, \bfphi^\top \bfx^s).
\end{aligned}
\end{equation}
The second term of the objective can be rewritten as follows,

\begin{equation}
\label{equ:term2}
\begin{aligned}
&C_2 \sum_{i=1}^{n_1}  \left \|\pi_i - \sum_{k\in \mathcal{N}^s_i} \omega_{ik}^s \pi_k \right \|_2^2\\
&=C_2 \sum_{i=1}^{n_1}  \left \|\bfa_i^\top \bfpi - \bfomega_i^\top \bfpi \right \|_2^2\\
&=C_2 \sum_{i=1}^{n_1}  \left \|\left (\bfa_i - \bfomega_i \right ) ^\top \bfpi \right \|_2^2\\
&=C_2 \sum_{i=1}^{n_1}  \bfpi^\top \left (\bfa_i - \bfomega_i \right )  \left (\bfa_i - \bfomega_i \right ) ^\top \bfpi\\
&=\bfpi^\top \left (C_2 \sum_{i=1}^{n_1}   \left (\bfa_i - \bfomega_i \right )  \left (\bfa_i - \bfomega_i \right ) ^\top  \right) \bfpi,
\end{aligned}
\end{equation}
where $\bfa_i \in \{1,0\}^{n_1}$ and its $i$-th element is one, while the other elements are zeros. $\bfomega_i \in \mathbb{R}^{n_1}$ and its $k$-th element is defined as follows,

\begin{equation}
\label{equ:omega}
\begin{aligned}
\bfomega_{ik}=
\left\{\begin{matrix}
\omega_{ik}, &if~k\in \mathcal{N}_i^s \\
0, &otherwise.
\end{matrix}\right.
\end{aligned}
\end{equation}
The third term of the objective function is rewritten as follows,

\begin{equation}
\label{equ:C3}
\begin{aligned}
&\frac{C_3 }{2} \left \| \frac{1}{n_1} \sum_{i=1}^{n_1} \Theta \bfx_i^s \pi_i -
\frac{1}{n_2} \sum_{j=1}^{n_2} \Theta \bfx_j^t \right \|_2^2\\
&=\frac{C_3 }{2} \left \| \Gamma \bfpi - \bfvartheta \right \|_2^2\\
&=\frac{C_3 }{2} Tr \left (( \Gamma \bfpi - \bfvartheta )^\top ( \Gamma \bfpi - \bfvartheta )\right )\\
&=\frac{C_3 }{2} \left ( \bfpi^\top \Gamma ^\top \Gamma \bfpi - 2 \bfvartheta^\top \Gamma \bfpi   + \bfvartheta^\top \bfvartheta
\right ),\\
&where~\Gamma = \left [\frac{1}{n_1} \Theta \bfx_1^s, \cdots, \frac{1}{n_1} \Theta \bfx_{n_1}^s \right ] \in \mathbb{R}^{r\times n_1},\\
&and ~\bfvartheta=\frac{1}{n_2} \sum_{j=1}^{n_2} \Theta \bfx_j^t\in \mathbb{R}^{n_1}.
\end{aligned}
\end{equation}
Substituting (\ref{equ:linear}), (\ref{equ:term2}), and (\ref{equ:C3}) back to (\ref{equ:minimpi}), we have

\begin{equation}
\label{equ:minimpi1}
\begin{aligned}
\min_{\bfpi}~&
\bftau^\top \bfpi+\bfpi^\top \left (C_2 \sum_{i=1}^{n_1}   \left (\bfa_i - \bfomega_i \right )  \left (\bfa_i - \bfomega_i \right ) ^\top  \right) \bfpi\\
&+\frac{C_3 }{2} \left ( \bfpi^\top \Gamma ^\top \Gamma \bfpi - 2 \bfvartheta^\top \Gamma \bfpi + \bfvartheta^\top \bfvartheta
\right )\\
s.t.~&\bf0 \leq\bfpi \leq \delta\bf1, ~and~\bfpi^\top \bf1 = n_1.
\end{aligned}
\end{equation}
This is a quadratic programming problem with linear constraints. We can solve this problem by active set algorithm.

\subsection{Iterative algorithm}

The iterative learning algorithm of our proposed method is summarized in Algorithm \ref{alg}.

\begin{algorithm}
\caption{Iterative learning algorithm of shared subspace and partially shared classifiers (SSPSC).}
\label{alg}
\begin{algorithmic}
\STATE \textbf{Input}: Source and target domain training sets, $\{(\bfx_1^s, y_1^s), \cdots, (\bfx_{n_1}^s, y_{n_1}^s)\}$ and $\{(\bfx_1^t,y_1^t), \cdots, (\bfx_{n_3}^t,y_{n_3}^t), \bfx_{n_3+1}^t,\cdots, \bfx_{n_2}^t\}$;
\STATE \textbf{Input}: Weight factors of different terms $C_1$, $C_2$ and $C_3$;
\STATE Initialize $\bfphi$, $\bfvarphi$, and $\bfpi$;
\REPEAT
\STATE Update $\Theta$ by solving (\ref{equ:minTheta1});
\STATE Update $\bfw$ by (\ref{equ:woptimal});
\STATE Update $\bfphi$ and $\bfvarphi$ by (\ref{equ:updateuv});
\STATE Update $\bfpi$ by solving (\ref{equ:minimpi1});
\UNTIL{Converge}

\end{algorithmic}
\end{algorithm}

\section{Experimental Results}
\label{sec:exp}

In this section, we conduct some experiments over five benchmark domain transfer learning data sets to evaluate the performances of the proposed method. We firstly test how the proposed method works with different values of term weights, $C_1$, $C_2$, and $C_3$. Then we test the classification performance of the proposed method by comparing it to different transfer learning methods, both in term of classification accuracy and running time.
%Finally, because the proposed method is an iterative algorithm, we are also interested in the convergence, and we test the convergence performance of the algorithm.

\subsection{Benchmark Data Sets}

In the experiments, we use five benchmark data sets. They are listed as follows.

\begin{itemize}
\item \textbf{Travel destination review data set} is a data set of reviews to several destinations of touring. In this data set, we have four Europe destinations of traveling, which are London, Rome, Paris, and Venice. For each destination, we collected 200 positive reviews and 200 negative reviews. We treat each travel destination as a domain, and each review as a data point. In the experiment, we randomly select one destination as a source domain, and select another destination as a target domain. To extract features from a review, we use the bag-of-words features.

\item \textbf{20-Newsgroup corpus data set} is a data set of newspaper documents. It contains documents of 20 classes. The classes are organized in a hierarchical structure. For a class, it usually have two or more sub-classes. For example, in the class of \emph{car}, there are two sub-classes, which are \emph{motorcycle} and \emph{auto}. To split this data set to source domain and target domain, for one class, we keep one sub-class in the source domain, while put the other sub-class to the target domain. We follow the splitting of source and target domain of NG14 data set of \cite{Chen20131284}. In this data set, there are 6 classes, and for each class, one sub-class is in the source domain, and another sub-class is in the target domain. For each domain, the number of data points is 2,400. The bag-of-word features of each document are used as original features.

\item \textbf{Amazon review data set} is a data set of reviews of products. It contains reviews of three types of products, which are books, DVD and Music. The reviews belongs to two classes, which are positive and negative. We treat the review of books as source domain, and that of DVD as target domain. For each domain, we have 2,000 positive reviews and 2,000 reviews. Again, we use the bag-of-words features as the features of reviews.

\item \textbf{GEMEP-FERA face expression data set} is a data of videos of faces. In this data set, we have the 87 face videos of 7 individuals. We treat the each individual as a domain, and each frame of the videos as a data point. We randomly select one individual as a source domain, and another individual as a target domain. The problem of classification is to classify a given frame to one of the 7 face expression classes.

\item \textbf{Spam email data set} is a set of emails of different individuals. In this data set, there are emails of three different individuals' inboxes, and we treat each individual as a domain. In each individual's inbox, there are 2,500 emails, and the emails are classified to two different classes, which are normal email and spam email. we also randomly choose one individual as a source domain, and another one as a target domain.
\end{itemize}

\subsection{Experiment Process}

In the experiments, we use the $10$-fold cross validation. We set all the source domain data points labeled data points, and use all of them in the training process. Moreover, we split the target domain set to ten folds. Each fold is used as a test set, and the other folds are combined and used as a training set. For the training set, we randomly choose a half of the data points and set them as labeled data points, and leave the remaining half as unlabeled. We use the source domain training set and the target domain training set to train the parameters of our model using the proposed algorithm, and then apply the trained model to the test set and evaluate the classification performances. For the multi-class classification problem, we extend the proposed binary classification model to multi-class classification by the one-vs-all strategy. For the data set with more than two domains, we use each domain as a target domain in turns, and randomly choose anther domain as a source domain. The accuracies of over all the target domains are averaged and reported as the final results.

\subsection{Sensitivity to Term Weights}

\begin{figure}
\includegraphics[width=0.7\textwidth]{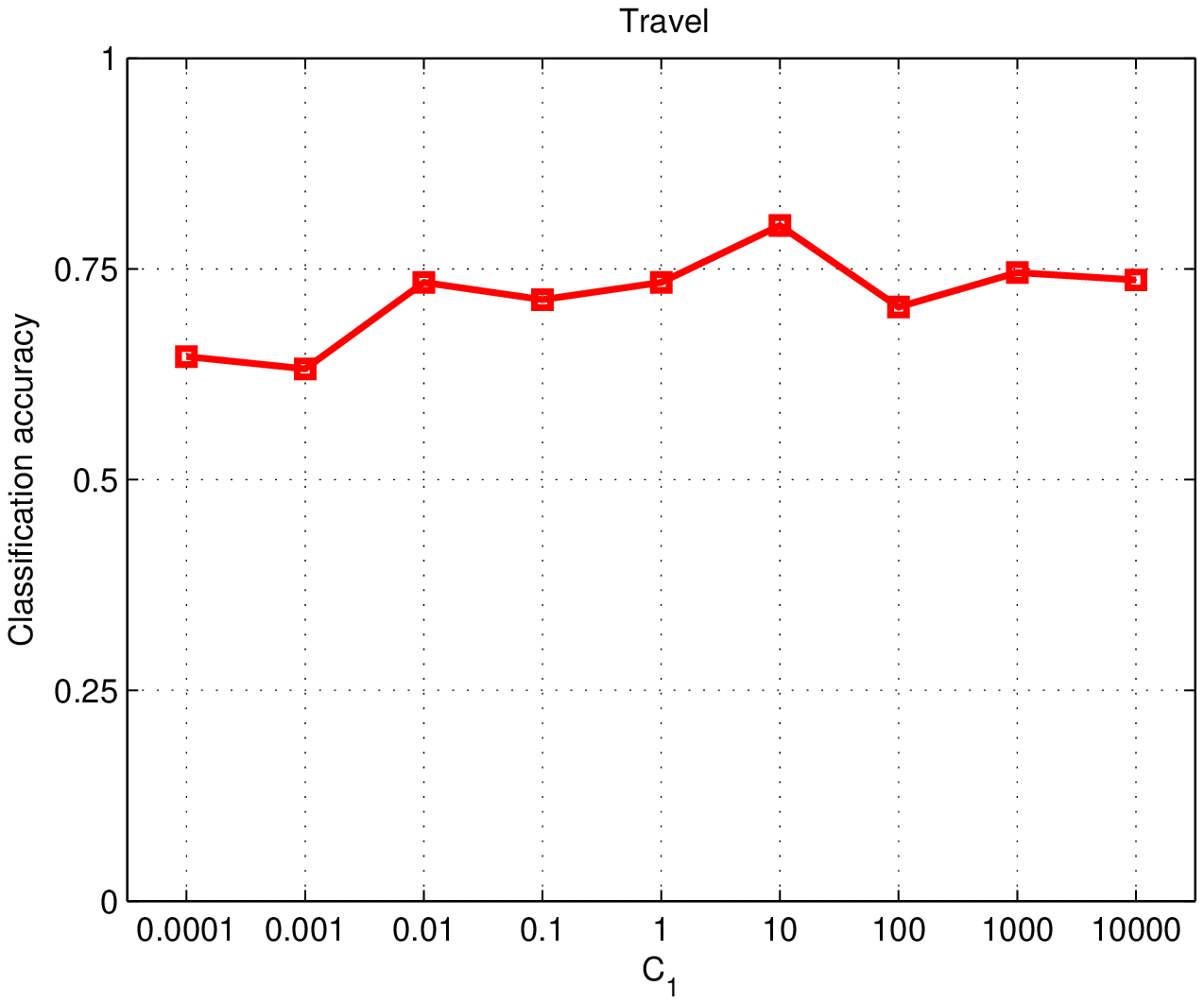}\\
\includegraphics[width=0.7\textwidth]{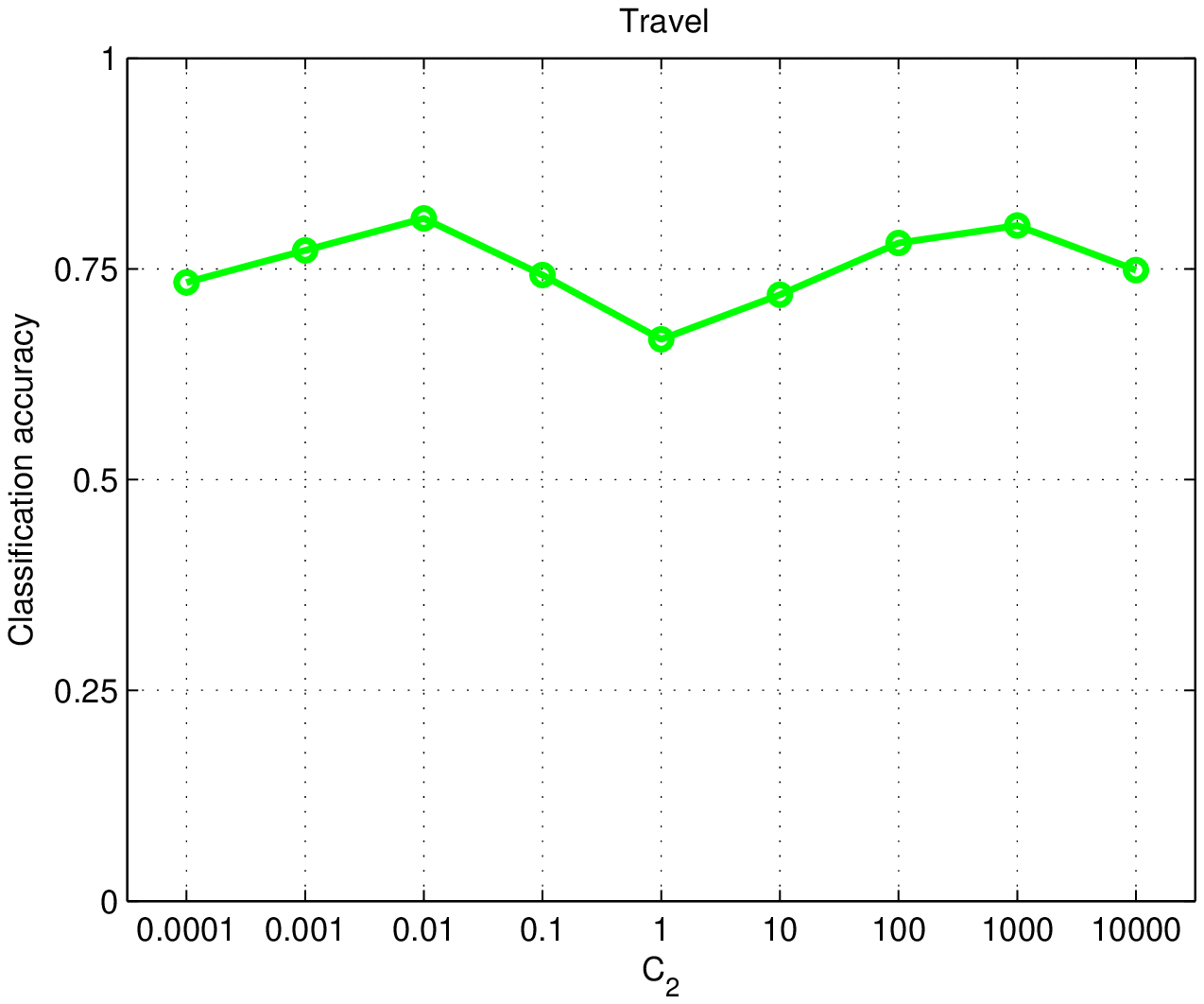}\\
\includegraphics[width=0.7\textwidth]{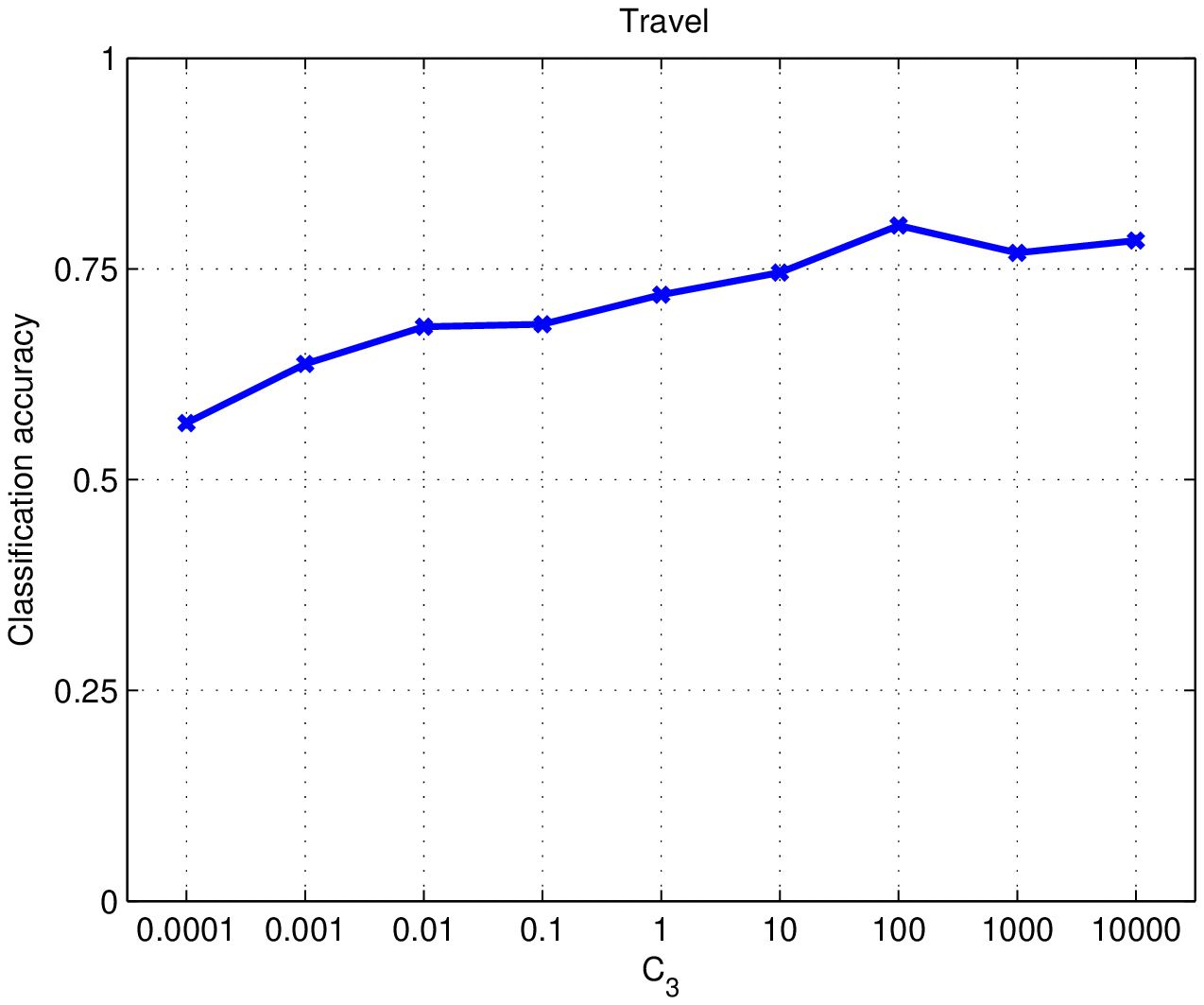}
\caption{Sensitivity curves to the term weights, $C_1$, $C_2$ and $C_3$ over Travel data set.}
\label{fig:sen}
\end{figure}

We study the sensitivity to the term weights of the objective function, $C_1$, $C_2$, and $C_3$. As an example, we use the data set of travel destination reviews. The accuracies of the proposed algorithm with different values of $C_1$, $C_2$, and $C_3$ are reported in Figure \ref{fig:sen}. From the figure, we can see that the proposed method is stable to weight $C_1$. The highest accuracy is obtained when $C_1$ is set to 10. For $C_2$, the accuracies are also stable to the changes of the values. The accuracies are around 0.75. However, for $C_3$, we have a clear trend that the accuracies are increasing with larger values of $C_3$.

\subsection{Comparison to Other Transfer Learning Methods}

We compare the proposed method to the methods listed in the section \ref{sec:existingworks}. The comparison is reported in terms of classification accuracy and running time.

\subsubsection{Classification Accuracy}

The classification accuracies of the compared methods over five benchmark data sets are reported in Table \ref{tab:acc}. The proposed method outperforms all the compared methods over four benchmark data sets. The only exception is the case of the face expression classification problem over GEMEP data set, where the method of Chu et al. \cite{Chu6619295} obtains slightly better performance than the proposed method. However, even in the experiments over GEMEP, the proposed method still has the second best performance. In the experiments over both Travel destination review data set and the 20-Newsgroup data set, the proposed method outperforms the other methods significantly.

\begin{table}
\centering
\caption{Classification accuracy of compared methods over benchmark data sets.}
\label{tab:acc}
\begin{tabular}{|l||r|r|r|r|r|}
\hline
Methods & Travel & 20-Newsgroup & Amazon & GEMEP & Spam \\\hline
\hline
Proposed                        &0.8015&0.6210&0.7812&0.6450&0.8641\\\hline
Chen et al. \cite{Chen20131284} &0.6841&0.5815&0.7621&0.6214&0.8514\\\hline
Chu et al. \cite{Chu6619295}    &0.7642&0.5471&0.7642&0.6715&0.8354\\\hline
Ma et al. \cite{ma2014knowledge}&0.7435&0.5164&0.7255&0.6358&0.8012\\\hline
Xiao and Guo \cite{Xiao201554}  &0.7033&0.5236&0.7462&0.6451&0.8294\\\hline
Li et al. \cite{Li20141134}     &0.7134&0.5615&0.7134&0.6154&0.8122\\\hline
\end{tabular}
\end{table}

\subsubsection{Running Time}

The running time of the training process of the compared algorithms over different benchmark data sets are reported in Table \ref{tab:time}. From the reported results, we can see that the proposed method has the least running time. Moreover, we can also see that the running time is also relevant to the size of the data set. For example, in the two smallest data set, Travel and GEMEP data sets, the running time is also shorter  than the running time over other data sets.

\begin{table}
\centering
\caption{Running time of compared methods over benchmark data sets (seconds).}
\label{tab:time}
\begin{tabular}{|l||r|r|r|r|r|}
\hline
Methods & Travel & 20-Newsgroup & Amazon & GEMEP & Spam \\\hline
\hline
Proposed                        &15.51&79.16&60.45&20.16&94.51\\\hline
Chen et al. \cite{Chen20131284} &28.45&91.66&86.21&25.18&120.43\\\hline
Chu et al. \cite{Chu6619295}    &18.41&86.38&65.77&23.64&100.67\\\hline
Ma et al. \cite{ma2014knowledge}&21.83&92.14&71.94&31.68&136.18\\\hline
Xiao and Guo \cite{Xiao201554}  &19.66&84.16&66.34&26.96&140.57\\\hline
Li et al. \cite{Li20141134}     &35.09&100.60&70.14&30.01&134.10\\\hline
\end{tabular}
\end{table}

\section{Conclusions}
\label{sec:conclu}

In this paper, we proposed a novel transfer learning method. The features of this work is listed as follows,

\begin{itemize}
\item Instead of learning a shared representation and classifier directly for both source and target domains, we proposed to learn shared subspaces and classifier, and then adapt it to source and target domains.

\item Instead of using the source domain data points equally to estimate the distribution of the source domain, we proposed to weight the source domain data points in the subspaces to match the distributions of the two domains.

\item We also proposed to regularize the weighting factors of the source domain data points and the classification responses of the target domain data points by the local reconstruction coefficients.
\end{itemize}
The minimization problem of our method is based on these features, and we solve it by an iterative algorithm. Experiments show its advantages over some other methods.

\section{Future Works}
\label{sec:future}

In the future, we will study extending the proposed method to extremely large data sets, i.e., big data. We have two strategies to change the proposed algorithm to scale to big data sets. The first strategy is to parallelize the algorithm. The big data set can be split to many small sub-sets and the algorithm can be parallelized to process these sub-sets simultaneously. The second strategy is to use the stochastic optimization method by using the data points one by one, not using all of them sententiously. We also will extend the proposed algorithm to various applications, such as computational mechanic \cite{wang2014computational,zhou2014biomarker,liu2013structure,peng2015modeling,xu2016mechanical,zhou2016mechanical}, %natural language processing, computer vision,
multimedia\cite{wang2014effective,lin2016multi,liu2015supervised,wang2015multiple,wang2015supervised}, nanotechnology \cite{liu2016optofluidic,liu2015electro,liu2014correlated,liu2013effect,chen2011dual}, etc. We will also consider use some other models to represent and construction the classifier, such as Bayesian network \cite{fan2014tightening,fan2014finding,fan2015improved}.

\section*{Acknowledgements}

This research was supported by National Natural Science Foundation (71173062, 71203047), Key Program for Science and Technology Research of Heilongjiang Province (GB14D201) and University Academic Innovation Team Construction Plan of Philosophy and Social Sciences in Heilongjiang Province (TD201203).

%%\bibliographystyle{spmpsci}
%%\bibliography{Part}

\end{document}